# Bridging Source and Target Domains via Link Prediction for Unsupervised Domain Adaptation on Graphs

Yilong Wang
The Pennsylvania State University
State College, USA
yvw576@psu.edu

Tianxiang Zhao
The Pennsylvania State University
State College, USA
tkz5084@psu.edu

Zongyu Wu
The Pennsylvania State University
State College, USA
zongyuwu@psu.edu

Suhang Wang
The Pennsylvania State University
State College, USA
szw494@psu.edu

## Abstract

Graph neural networks (GNNs) have shown great ability for node classification on graphs. However, the success of GNNs relies on abundant labeled data, while obtaining high-quality labels is costly and challenging, especially for newly emerging domains. Hence, unsupervised domain adaptation (UDA), which trains a classifier on the labeled source graph and adapts it to the unlabeled target graph, is attracting increasing attention. Various approaches have been proposed to alleviate the distribution shift between the source and target graphs to facilitate the classifier adaptation. However, most of them simply adopt existing UDA techniques developed for independent and identically distributed data to gain domain-invariant node embeddings for graphs, which do not fully consider the graph structure and message-passing mechanism of GNNs during the adaptation and will fail when label distribution shift exists among domains. In this paper, we proposed a novel framework that adopts link prediction to connect nodes between source and target graphs, which can facilitate message-passing between the source and target graphs and augment the target nodes to have "in-distribution" neighborhoods with the source domain. This strategy modified the target graph on the input level to reduce its deviation from the source domain in the embedding space and is insensitive to disproportional label distributions across domains. To prevent the loss of discriminative information in the target graph, we further design a novel identity-preserving learning objective, which guides the learning of the edge insertion module together with reconstruction and adaptation losses. Experimental results on real-world datasets demonstrate the effectiveness of our framework.

## 1 Introduction

Graphs are pervasive in the real world, serving as fundamental structures in numerous domains such as social networks [38, 52], biological systems [25, 28], and transportation networks [11, 51]. Graph Neural Networks (GNNs) [3] have shown great power for representation learning on graphs with their iterative message-passing mechanism. Despite their success in various graph-learning tasks [13, 15, 32, 48, 49, 56], training GNN classifiers generally requires abundant high-quality labels; while obtaining labels is usually time-consuming, costly, and sometimes requires strong domain knowledge, which drastically impedes the adoption of GNNs [12]. As a result, transferring knowledge from a label-rich source domain to the target domain has been a trending topic.

Unsupervised domain adaptation (UDA) [42] has gained attention as a viable approach to address domain shifts. Typically, UDA trains a GNN classifier on a labeled source domain and adapts it to an unlabeled target domain. A key challenge lies in the inherent distribution shifts between domains, which often render the source-trained classifier ineffective in the target domain.

To address the distribution shift, one major idea is to align the embeddings of source and target nodes in a shared space, thus enabling a classifier trained on the source graph to generalize effectively to the target graph. Representative techniques such as adversarial training with domain discriminators [7, 34, 39] and discriminative distance regularization [23, 35] have been extensively explored in images domain and subsequently adapted for graph data [4, 31, 42, 53]. However, these methods exhibit three major drawbacks: (i) They primarily focus on learning domain-invariant node representations based only on the last layer output of the graph encoder, but the selection of node pairs across domains for the alignment is intrinsically challenging. (ii) These methods operate under the assumption that different domains have similar class distribution, and will fail when the disproportional label distribution occurs among domains. A detailed analysis of this issue is provided in Section 3.2. (iii) These methods do not fully leverage the unique properties of graph structures and the intrinsic message-passing mechanisms of GNNs.

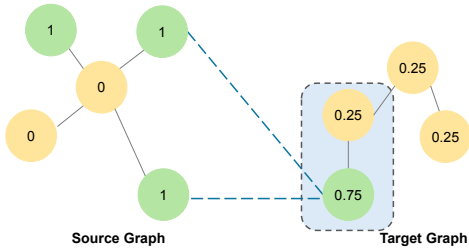

Figure 1: This figure depicts a simplified cross-domain graph for node classification, with pale blue lines indicating potential cross-domain edges.

To alleviate the above issues, we observe that (i) unlike images which are independently and identically distributed (i.i.d.), graphs inherently exhibit interdependencies among nodes through edges; and (ii) current GNNs extensively utilize the message-passing mechanism to propagate information through these connections and revise local neighborhood distributions. This unique message-passing mechanism on graphs inspires us to explore a new perspective addressing distribution shifts on graph datasets: *whether we can insert cross-domain edges as bridges and then propagate messages from source nodes to the target ones*. Figure 1 illustrates the potential of connecting two domains to mitigate domain distribution shifts and facilitate domain adaptation, where the number in each node denotes the node feature and the color denotes the class label. Due to the label and feature distribution shifts among domains, if nodes in the target domain merely aggregate information from in-domain neighbors, their embeddings become smoothed and less distinct, leading to challenges in classification. For example, with merely in-domain message passing, the embedding of the green node in the bounded area in the target domain will move toward the embedding space of the yellow class. However, with the injection of cross-domain edges connecting similar node pairs, the green node in the target domain could receive information from the labeled source domain and then push the embedding closer to the distribution center in the source domain. Then, a model trained in the source domain with supervised information can easily classify the target green node into the correct class. Another insight shown in figure 1 is that if edges are connected between cross-domain node pairs with similar embeddings, the model has a higher likelihood of incorporating intra-class edges. By augmenting intra-class edges, nodes of the same class from different domains can directly exchange information, leading to the conditional distribution $P_t(E|y)$ in the target domain closely approximating the conditional distribution $P_s(E|y)$ in the source domain, where $E$ denotes the node representation and $y$ denotes the label. This alignment of the conditional shift ensures that the operation of adding edges is unaffected by the label shift.

Therefore, in this work, we pioneer a distinctive perspective for addressing the graph domain adaptation problem by incorporating cross-domain edges to facilitate message passing and reduce distribution discrepancy between the source and target graphs. The insertion of cross-domain edges poses two main challenges: (i) How to effectively and efficiently add realistic cross-domain links? (ii) The added links might smooth source and target node representations. How to keep class discriminativeness of the target domain? Addressing these challenges, we propose a novel framework: Cross-domain Message Passing Graph Neural Network (CMPGNN), which dynamically inserts cross-domain edges with adaptive weight for adaptation. With established links as bridges, the message-passing mechanism enables information-deprived target nodes to receive input from nodes in the source domain. This information flow through cross-domain edges encourages the embeddings of the target graph to move close to the source domain, ensuring that the target data representations are not out-of-distribution (OOD). Moreover, we introduce a mutual information loss function to guarantee that, while benefiting from the source domain's information, the unique characteristics of target nodes are preserved. In experiments, we examined our model's performance under various label shift scenarios on large and complex datasets. The performance of our model sets a new SOTA in the field of graph domain adaptation. The main contributions of this work are:

- We introduce a novel perspective on the graph UDA problem by leveraging cross-domain edges to enhance knowledge transfer across domains, which directly aligns different domains on the input level and does not require the nonexistence of label shifts.
- We propose a new framework integrating a link predictor and a novel mutual information loss to inject cross-domain edges, which can reduce domain discrepancies and preserve class discriminativeness.
- Through extensive experiments on both well-established benchmarks and newly created ones, we demonstrate the effectiveness of our model in addressing domain shifts for graph data.

## 2 Relate Work

**Graph Neural Networks**. GNNs leverage graph structures to perform convolution operations for effective feature extraction. They are primarily divided into spectral-based [3, 15–17, 57] and spatial-based approaches [1, 8, 9, 18, 26, 36, 43, 54, 55]. Spectral GNNs operate in the spectral domain using graph Laplacian filters, while spatial GNNs aggregate information from neighboring nodes. Notable advances include GAT [36], which adopts attention mechanisms to aggregate neighbors. To address label scarcity, methods in semi-supervised [15, 45] and self-supervised learning [10, 37] have been explored.

**Graph Domain Adaptation**. Domain adaptation aims to mitigate the impact of distribution shifts between different domains on the model's transferability, thereby facilitating the application of the model in the target domain. While some previous works [2, 44] in the CV field connect similar image instances to construct subgraphs for domain adaptation, they still primarily address domain adaptation on i.i.d. data settings. Thus, their methodology is fundamentally different from graph domain adaptation, which is in non-i.i.d. environments. Numerous methods addressing domain shifts on graphs have been proposed. Struc-RW [21] re-weights the source adjacent matrix to mimic the connection pattern in the target domain to mitigate structure shift. Pair-Align [22] follows the re-weight strategy and considers label weights difference to adjust the classification loss. However, these re-weight-based methods modify the source domain's connection pattern and capitalizes only on partial information from the source domain, which impairs the capabilities learned by the model in the source domain. Other methods [42, 53] in GDA mainly focus on mitigating the influence of covariate shifts among domains. DANE [53] and ACDNE [30] introduce the technique of adversarial training on domain classifiers, which aligns the embedding to obtain domain-invariant representations. UDAGCN [42] expands the scope by considering the global level message passing. SGDA [29] adds a pseudo-label selection process to mitigate label scarcity. ASN [50] adds domain-private encoders to learn domain-specific information. A2GCN [19] modifies the structure of the propagation layers to decrease the influence of the source domain. GRADE [41] introduces a novel Graph Subtree Discrepancy metric to measure the distance between different domains in graph structures. Similarly, SpecReg [46] proposes the utilization of graph spectral regularization to tackle graph domain adaptation issues. While these methods have made significant progress in aligning embeddings from different domains to minimize covariate shifts, they encounter several challenges. Firstly, these methods predominantly align the outputs of the final layer of graph encoders, and [21] have theoretically demonstrated that aligning only the last layer embeddings is insufficient for graph data. Secondly, they often fail to fully leverage the unique message-passing mechanism inherent in graph neural networks. Lastly, they typically assume that there is no label shift across domains, an assumption that may not be valid in real-world scenarios.

Our work is inherently different from existing methods in that we explore a novel perspective: adding cross-domain edges to facilitate the cross-domain message passing and thereby mitigate the distribution shifts. In our model, the modification happens on the input level and the inserted cross-domain edges will facilitate the entire process of the message passing. Besides, our model does not conduct explicit alignments and, hence does not rely upon the assumption of invariant label distributions $p(Y)$.

## 3 Preliminary
### 3.1 Notations and Problem Definition

Let $\mathcal{G} = (\mathbb{V}, A, X, Y)$ denote an attributed graph, where $\mathbb{V} = \{v_1, ..., v_N\}$ is the set of $N$ nodes, and $A \in \mathbb{R}^{N \times N}$ is the adjacency matrix of $\mathcal{G}$, where $A_{i,j} = 1$ if node $v_i$ and node $v_j$ are connected, otherwise $A_{i,j} = 0$. $X \in \mathbb{R}^{N \times d}$ is the node attribute matrix with $X[i,:] \in \mathbb{R}^{1 \times d}$ being node attribute vector for node $v_i$ and $Y \in \mathbb{R}^{N \times C}$ denotes the label matrix where $C$ is the number of categories.

Let $\mathcal{G}^S = (\mathbb{V}^S, A^S, X^S, Y^S)$ be the labeled source graph, where $A^S, X^S$, and $Y^S$ are the adjacency matrix, node feature matrix, and label matrix, respectively. Similarly, let $\mathcal{G}^T = (\mathbb{V}^T, A^T, X^T)$ be a fully unlabeled target graph. Given $\mathcal{G}^S$ and $\mathcal{G}^T$, our goal is to employ both the source and target graphs to train a model for node classification on the unlabeled target graph $\mathcal{G}^T$.

Our model involves inserting cross-domain edges to connect graphs in two domains, resulting in a combined graph defined as: $\mathcal{G}^C = (\mathbb{V}^C, A^C, X^C, Y^C)$. This graph encompasses all nodes and edges from both the source and target domains. Initially, the adjacent matrix $A^C \in \mathbb{R}^{(N^S+N^T) \times (N^S+N^T)}$ is $\begin{bmatrix} A^S & 0 \\ 0 & A^T \end{bmatrix}$. During the training phase, our model dynamically incorporates cross-domain edges into the combined graph, refining the adjacency matrix $A^C$ with adaptively learned edge weights.

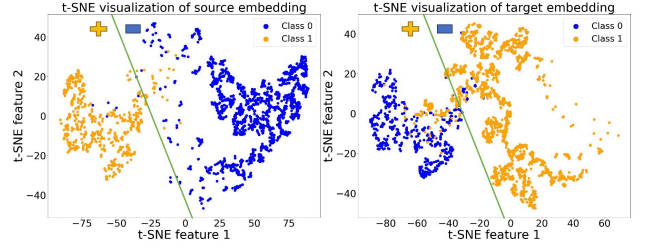

**Figure 2: Visualization of embeddings on the generated dataset with label shift: the left shows source domain embeddings, the right shows target domain embeddings, with the green line representing the source domain's classification boundary.**

### 3.2 Preliminary Analysis

In this subsection, we conduct a preliminary experiment to show the influence of the label shift on previous alignment methods and why cross-domain edge insertion can alleviate this issue.

To control label shift, we use the Conditional Stochastic Block Model (CSBM) [5] to construct source and target graphs with two node classes, Class A and Class B. Edges within each domain are sampled from the same distribution, with $p_{\text{intra}} = 0.08$ and $p_{\text{inter}} = 0.02$, eliminating structural shifts. Node features are drawn from class-specific normal distributions, e.g., Class A in the source domain follows $\mathcal{N}(\mu_A^S, \sigma_A^{S^2})$, while Class B in the target domain follows $\mathcal{N}(\mu_B^T, \sigma_B^{T^2})$, with slight feature distribution changes to simulate covariate shift. A class ratio difference (3:1 in the source vs. 1:3 in the target) exemplifies label shift.

Upon these graphs, we apply an existing alignment method, UDAGCN [42], which integrates a classification loss on the source domain with an adversarially trained domain loss. We then use T-SNE to visualize the learned node embedding. Figure 2 shows the node embedding for both domains when maximizing the domain loss during training. The visualizations show that: The blue dots in the target domain (right figure) are mismatched to the embedding space of the yellow dots in the source domain (left figure), and the yellow dots in the target domain are mismatched to the embedding space of the blue dots in the source domain. This is because the alignment methods only aim to align the marginal distributions of the source domain $P_S(H)$ and the target domain $P_T(H)$ to deceive the discriminator. Given that the joint distribution $P(H)$ can be decomposed as $P(H) = \sum_y P(H|y) P(y)$, when the label shift $P_S(y) \neq P_T(y)$ exists among domains, achieving the alignment of $P_S(H) = P_T(H)$ does not guarantee that the conditional distributions $P_S(H|y)$ and $P_T(H|y)$ are also aligned. Consequently, the classifier, which is trained on the source domain, may still suffer from degraded performance when applied to the target domain due to this conditional distribution mismatch.

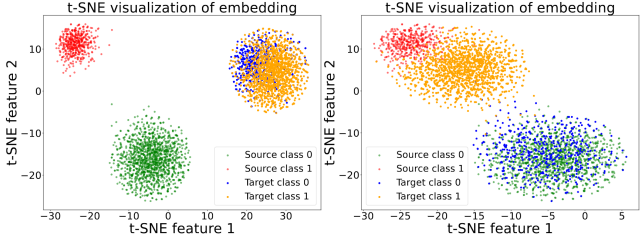

Figure 3: Visualization of embeddings on the generated dataset before and after inserting cross-domain edges. The left figure illustrates the embeddings from the source and target domains prior to edge insertion, and the right figure displays the embeddings post-insertion. Different colors denote distinct classes.

We further explore the influence of adding cross-domain edges on representation learning. We insert edges based on feature similarity. Specifically, if the similarity between a node $v_i \in \mathcal{G}^S$ and a node in $v_j \in \mathcal{G}^T$ is larger than threshold $t$, we then add an edge between node pair $(v_i, v_j)$ with edge weight as 1. Then we apply a 1-hop message passing, i.e., $\mathbf{AX}$, on the modified graph to examine whether the inserted edges can reduce distribution shifts between the source and target domains. Figure 3 depicts the embedding of the source and target domain, before and after the insertion of cross-domain links. The visualizations in Figure 3 demonstrate that the introduction of cross-domain edges prompts the embeddings of each class in the target domain to gravitate toward the embedding centers of the corresponding classes in the source domain. This realignment mitigates the conditional shift among domains and enhances the separation between classes within the target domain, potentially improving the efficacy of classifiers trained on source domain data when applied to the target domain. This preliminary example underscores the possibility of mitigating conditional shifts by inserting cross-domain edges under the existence of label shifts.

## 4 Methodology

To mitigate the distribution shifts between source and target graphs, we propose a novel approach to instigate cross-domain edges as bridges to facilitate the message passing from the source to the target domain. This strategy aims to yield target instance embeddings that are not only in-distribution but also preserve class discriminativity. Figure 4 illustrates the proposed CMPGNN, which is composed of three primary components: a graph encoder $f_E$, a link predictor $f_P$, and a classifier $f_c$. The graph encoder takes graphs as input to learn node representations that capture node features and local neighborhood structure. The link predictor takes the learned representation to add links between source and target graphs, aiming to facilitate cross-domain message passing. The classifier takes the node representation as input to predict node labels. During training, beyond the classification loss on $\mathcal{G}^S$ and entropy loss on the target domain, we introduce two novel signals: a link prediction loss to generate realistic edges and a mutual information loss to encourage the preservation of discriminative node features. Next, we introduce the details of each component.

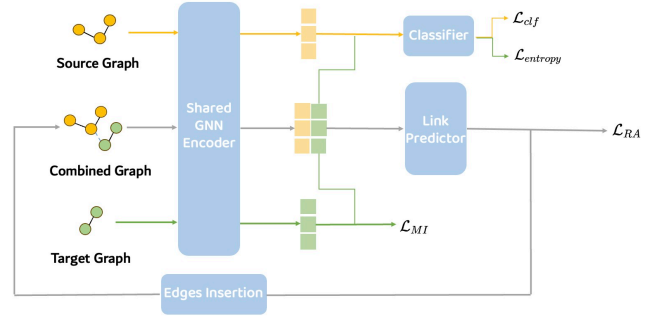

Figure 4: An illustration of the proposed CMPGNN

### 4.1 Node Representation Learning

Both the cross-domain edge prediction and node classification on different domains rely on high-quality node representations. In this work, we adopt GNNs as the graph encoder $f_E$. GNNs are highly effective in learning node representations as they capture both local graph structures and neighborhood information. To facilitate alignment of node representations between source and target domains, we adopt a shared-weight multi-layer GNN encoder $f_E$ to extract feature information on both $\mathcal{G}^S$ and $\mathcal{G}^T$. Specifically, the encoder takes node features and the adjacency matrix as inputs and outputs a node embedding matrix $H$ for each graph as

$$H^S = f_E(A^S, X^S), \quad H^T = f_E(A^T, X^T), \quad H^C = f_E(A^C, X^C) \quad (1)$$

where $H^S$, $H^T$, and $H^C$ are node representation matrices for source graph $\mathcal{G}^S$, target graph $\mathcal{G}^T$, and combined graph $\mathcal{G}^C$, respectively. Various GNNs can be used as $f_E$ such as GCN [15] and GIN [43].

### 4.2 Cross-Domain Link Predictor

The shared encoder $f_E$ maps nodes from disparate domains into a unified embedding space, on top of which we can model the distribution of cross-domain edges. The primary concern of the link predictor is how to find feasible and realistic edges across domains. Besides, those newly added edges should not dominate the target node representation, which would lead to the loss of discriminative information. Moreover, link prediction operations are typically time-consuming, raising the critical issue of balancing efficiency and effectiveness. Addressing these challenges, we design a surrogate link prediction objective, propose a mutual-information-based loss to preserve node-wise information and adopt a quick filtering strategy on edge candidates to reduce the computation cost. Next, we give the details.

*4.2.1 Link Predictor Structure and Objective Function.* We employ a simple Multi-Layer Perception (MLP) $f_P$ to predict the links between domains. The MLP accepts node embeddings $H^C$ generated by the graph encoder $f_E$ from the combined graph, maps them into a new representation space $Z^C = f_P(H^C)$, and models the probability of edge existence between a pair of node $v_i$ and $v_j$ as:

$$S_{ij} = \sigma(\mathbf{z}_i^\top \mathbf{z}_j) \quad (2)$$

where $\sigma(\cdot)$ is the sigmoid activation function. $\mathbf{z}_i$ and $\mathbf{z}_j$ represent the embeddings of nodes $v_i$ and $v_j$ in the combined graph respectively. The computed probability score $S_{ij}$ can also be interpreted as the edge weight, indicating the strength of potential cross-domain links between these nodes.

To produce cross-domain edges that are realistic and "in-distribution" for the learned model, we propose to train the link predictor by reconstructing edges from the source graph. This reconstruction loss encourages similar link prediction embeddings among connected node pairs and different ones among disconnected ones. Given the sparsity of adjacency matrices in real-world networks, we employ a negative sampling strategy to counterbalance the over-representation of disconnected node pairs in the loss function, which, despite their numerical prevalence, provide less informative signals. The link prediction loss is given as:

$$\mathcal{L}_{RA} = \mathbb{E}_{i \in \mathbb{V}^C} \frac{1}{|N(i)|} \sum_{j \in N(i)} \left[ (S_{ij} - 1)^2 + \mathbb{E}_{k \in Neg_{(i,j)}} (S_{ik} - 0)^2 \right] \quad (3)$$

where $N(i)$ denotes the neighborhood set of node $v_i$ within the combined graph $\mathcal{G}^C$. For each connected node pair $(v_i, v_j)$, negative sampling $Neg(i, j)$ is utilized to select non-neighbor nodes randomly during each training epoch. This strategy is applied across all node pairs in $\mathcal{G}^C$, yielding a sample encompassing both intra-domain and cross-domain node pairs.

*4.2.2 Cross-domain Edges Insertion.* The learned edge predictor can be directly applied to generate realistic cross-domain connections in the shared representation space. However, it is not desirable to establish an excessive number of cross-domain edges, which will introduce too much additional information and change the distribution of the target graph dramatically. Besides, directly applying the learned edge predictor for all cross-domain node pairs is time-consuming, as the time complexity would be $O(N^S N^T)$ at each inference step, where $N^S$ and $N^T$ are the numbers of nodes in the source and target domains, respectively. Addressing this problem, we propose to accelerate by selecting a smaller candidate set so that links are more likely to be established and only predict links for the candidate set. Specifically, before initiating the training process, for each node $v_i$ in the target domain $\mathbb{V}^T$, we compute its embedding similarity of the pre-trained encoder with every node $v_j$ in the source domain $\mathbb{V}^S$. We then select the top-$K$ most similar nodes from the source domain to construct a candidate set $\mathbb{C}^i$ for each target node $v_i$. During experiments, we found that the performance on the target domain is stable relative to the size of the candidate set, and the candidate size $K$ is fixed at 50 for all datasets. The candidate set construction operation is in $O(N^S N^T)$, and we only conduct it once before training begins. Inside each training epoch, our model only calculates the similarity score $S$ between each target domain node and its corresponding candidate set. This strategic refinement reduces the time complexity of modifying the combined graph to $O(N^T)$, enhancing the feasibility of the procedure for large graphs.

During each training epoch, we identify and incorporate edges between source and target nodes exhibiting high connection probability into the combined graph with adaptive weights. Specifically, for each node $v_i \in \mathbb{V}^T$, and $v_j \in \mathbb{C}^i$, if the predicted connection probability $S_{ij}$ is larger than the threshold **t**, then we add the new edge $(v_i, v_j)$ into the modified combined graph $\hat{\mathcal{G}}^C$ with edge weight $S_{ij}$. The modified combined graph $\hat{\mathcal{G}}^C$ is defined as:

$$\hat{\mathcal{G}}^C = (\mathbb{V}^C, \hat{A}^C, X^C, Y^C) \quad (4)$$

where $\hat{A}^C$ is the updated adjacent matrix and has value as:

$$\hat{A}_{ij}^C = \begin{cases} 1 & \text{if } A_{ij}^C = 1; \\ S_{ij} & \text{else if } S_{ij} > t, v_i \in \mathbb{V}^T \text{ and } v_j \in \mathbb{C}^i; \\ 0 & \text{else,} \end{cases} \quad (5)$$

The threshold $t$ is fixed as 0.94 in the experiment. The updated graph $\hat{\mathcal{G}}^C$ will replace the previous one and serve as the input for the ensuing epoch.

The link insertion process establishes connections between similar cross-domain node pairs. As nodes of the same class across domains are expected to exhibit similar feature representations and connectivity patterns, our model inherently promotes the formation of cross-domain, intra-class edges. These edges act as bridges during message passing, enabling nodes in the target domain to directly receive and integrate information from same-class nodes in the source domain, which helps reduce class-wise embedding disparity between domains and mitigates the conditional shift $P(H|y)$.

### 4.3 Preserving Discriminativity with MI Loss

During the training procedure, messages propagated on cross-domain edges facilitate the representation of nodes in $\mathcal{G}^T$ to be more 'in-distribution' with nodes from the source domain, promoting the transferability of learned source models to work on them. However, cross-domain links also introduce noise into the target domain and bring a high risk of losing discriminative information for each target node and causing over-smoothing among domains. For example, there is one trivial solution for the cross-domain edges insertion process that all target nodes are connected to several informative nodes (like those with high degree and homophily ratio [24]) in the source domain and then are classified as the same class. What we expect is that, with the cross-domain edges as bridges, nodes in the target domain could receive information from the source domain while at the same time preserving their intrinsic uniqueness. To address this, we propose a novel mutual information (MI) loss $\mathcal{L}_{MI}$ between the combined graph and the target graph. This mutual information loss is designed to encourage semantic consistency during the process of adding cross-domain edges and preserve individual information of each target node by contrasting them with each other. The mutual information loss $\mathcal{L}_{MI}$ is:

$$\mathcal{L}_{MI} = \frac{-1}{N_T} \sum_{i=1}^{N_T} \log \frac{e^{(h_i^C, h_i^T)/\tau}}{e^{(h_i^C, h_i^T)/\tau} + \mathbb{E}_{k \in Neg_{(i)}} e^{(h_i^C, h_k^T)/\tau}}, \quad (6)$$

where $h_i^C$, and $h_i^T$ are different embeddings of the same node $v_i$ on the combined and target graph respectively. $e^{(\cdot, \cdot)}$ represents the exponential of the dot product. Negative sampling $Neg(i)$ is utilized to select non-self nodes to build negative samples for each training epoch. The temperature $\tau$ is fixed as 0.3 in the experiment.

### 4.4 Full Adaptation Objective

With cross-domain edges inserted into the combined graph, representations of target nodes will be more "in-distribution" and improve the transferability of the learned source model to work on them. In this part, we will introduce our objective to adapt the source classifier, $f_c$, to work on the shared embedding space and map node representation to predicted class labels. The classifier $f_c$ is a one-layer perception to predict the label $\hat{Y}$. We implement a supervised node classification loss, $\mathcal{L}_{clf}$ on source graph $\mathcal{G}^S$, and an unsupervised classification entropy loss $\mathcal{L}_{entropy}$ on the combined graph $\mathcal{G}^C$. Details will be introduced in subsequent parts.

*4.4.1 Classification Loss on Source Graph.* The source classification Loss $\mathcal{L}_{clf}(f_c(H^S), Y^S)$ is to minimize the cross-entropy loss for the labeled data in the source domain. The classification loss is:

$$\mathcal{L}_{clf}(f_c(H^S), Y^S) = -\frac{1}{N_S} \sum_{i=1}^{N_S} \mathbf{y}_i \log(\hat{\mathbf{y}}_i) \quad (7)$$

where $\mathbf{y}_i$ denotes the label of the i-th node in the source domain, $\hat{\mathbf{y}}_i$ are the classification prediction for the *i*-th source labeled node $v_i$, respectively. This classification loss will guide the learning of a precise mapping between the representation space and node labels. Notably, we do not include newly added edges in this supervised edge prediction to prevent the introduction of noises, especially at early learning stages.

*4.4.2 Entropy Loss on Target Graph.* For the nodes in the target domain, since we do not have the label information, we use the entropy loss $\mathcal{L}_{entropy}(f_c(H_T^C))$ to guide the training procedure. The entropy loss for nodes in the target domain is defined as:

$$\mathcal{L}_{entropy}(f_c(H_T^C)) = -\frac{1}{N^T} \sum_{i=1}^{N^T} \hat{\mathbf{y}}_i \log(\hat{\mathbf{y}}_i) \quad (8)$$

where $H_T^C = \{\mathbf{h}_i^C | v_i \in \mathbb{V}^T\}$ is the embedding of nodes in the target domain and aggregated on the combined graph. This entropy loss will reduce the ambiguity of model predictions on the target nodes, encouraging $f_c$ to give a clearer class prediction for them.

## 4.5 Final Objective Function

The final objective function contains four parts: the reconstruction loss $\mathcal{L}_{RA}$ to guide the cross-domain edge insertion process; the mutual information loss $\mathcal{L}_{MI}$ to preserve domain discriminative information, $\mathcal{L}_{clf}$ and $\mathcal{L}_{entropy}$ as the classification loss to learn the mapping function from the embedding to label space. The final loss function of our CMPGNN model is given as:

$$\mathcal{L} = \min_{\theta_E, \theta_c, \theta_P} \mathcal{L}_{clf} + \lambda_1 \mathcal{L}_{entropy} + \lambda_2 \mathcal{L}_{RA} + \lambda_3 \mathcal{L}_{MI} \quad (9)$$

where $\theta_E$, $\theta_P$, and $\theta_c$ are the parameter in the graph encoder $f_E$, the link predictor $f_P$, and the classifier $f_c$, respectively. $\lambda_1$, $\lambda_2$, and $\lambda_3$ are hyper-parameter used to balance the contribution of each part. $f_e$, $f_P$, and $f_c$ are jointly trained together with Eq. 9.

The whole training algorithm and time complexity analysis of our CMPGNN model are in the Appendix B.

## 5 Experiment

In this section, we conduct experiments to evaluate the effectiveness of the proposed framework. In particular, we aim to answer the following questions: (i) **RQ1**: How does the CMPGNN model perform for node classification tasks across different domains? (ii) **RQ2**: What properties do the edges predicted by CMPGNN between domains exhibit? (iii) **RQ3**: Whether the Mutual Information Loss term preserves domain-discriminative information? (iv) **RQ4**: How does CMPGNN cope with label shifts between the domains? Is the model able to alleviate conditional shift under label shift setting?

### 5.1 Experimental Settings

For a comprehensive evaluation, we benchmark our CMPGNN model against the following baseline methods, including the latest state-of-the-art domain adaptation techniques in graph domains: **GCN** [15], **GRADE** [41], **UDAGCN** [42], **ASN** [50], **StrucRW** [21], **SpecReg** [46], **A2GCN** [19]. We test these models on three real-world benchmarks: **ArnetMiner** [33], **Blog** [47], and **Twitch** [19], across 12 domain adaptation scenarios. Detailed descriptions of the baseline implementations and datasets can be found in Appendix A.

### 5.2 Node Classification Results

In addressing **RQ1**, we compared the performance of CMPGNN with baseline models across three real-world scenarios. We began by conducting experiments on the Citation and Blog datasets under two distinct conditions: (i) a fully-supervised setting where the source graph is entirely labeled, and (ii) a semi-supervised setting where only 10% of the nodes per class in the source graph have labels. In both scenarios, the target domain remains unlabeled. Each experiment was conducted five times, and we report the average results along with the standard deviation. Table 1 shows the result for fully-supervised setting and result under semi-supervised setting is shown in Appendix C. We then performed experiments on the Twitch dataset under a fully-supervised setting. The results are presented in Table 2. The result of all baseline models in Table 2 is inherited from work [20]. The AUROC score is used to evaluate model performance, given the significant label shift across domains. Our key observations are as follows:

- CMPGNN consistently outperforms all baselines across all cross-domain tasks in every experimental setting, underscoring the effectiveness of the proposed cross-domain edge insertion mechanism in mitigating distributional shifts.
- The performance differences between the semi-supervised and fully-supervised settings are minimal across all methods, suggesting that increasing labeled data in the source domain alone is insufficient to address domain shifts. This highlights the need for strategies that directly confront domain shifts to enhance model generalizability.
- In the presence of label shift, previous alignment methods often fail to significantly improve performance compared to the basic GCN model. However, our model maintains high performance even when label shifts occur.

### 5.3 Ablation Study

To elucidate the contributions of individual components to the overall efficacy of our proposed model, we conducted an exhaustive ablation study. This study examines three variants of our CMPGNN model: (i) **GCN+DA**: The Graph Encoder trained with classification loss on the source domain and entropy loss on the target domain; (ii) **Random Link**: The model omits the Link Predictor, randomly connecting edge pairs with higher similarity than the same threshold $t$ used in our model; and (iii) **Ours w/o MI**: The CMPGNN model excluding the Mutual Information Loss $\mathcal{L}_{MI}$ term. Performances of these variants compared with our CMPGNN on $ACM_L \rightarrow DBLP_L$ and $DBLP_L \rightarrow ACM_L$ tasks are depicted in Figure 5. The conclusions derived from Figure 5 are twofold: (i) The performance of Random Link is similar to that of GCN + DA, underscoring the critical role of the link predictor. This finding suggests that discerning informative cross-domain connections is critical for the success of CMPGNN; and (ii) The full CMPGNN model surpasses CMPGNN$_{w/o\ MI}$ in both experimental setups, highlighting the importance of mutual information loss in preserving class discriminativeness across domains.

Table 1: Classification Accuracy Under Fully-Supervised-Setting

| Methods | ACM $\to$ DBLP | DBLP $\to$ ACM | Blog1 $\to$ Blog2 | Blog2 $\to$ Blog1 | $ACM_L \to DBLP_L$ | $DBLP_L \to ACM_L$ |
|---|---|---|---|---|---|---|
| GCN | 76.01 ± 0.84 | 73.01 ± 0.74 | 56.97 ± 0.68 | 56.47 ± 0.54 | 79.01 ± 0.54 | 82.91 ± 0.71 |
| UDAGCN | 80.08 ± 0.88 | 75.55 ± 0.31 | 61.87 ± 0.54 | 57.67 ± 0.85 | 82.46 ± 0.34 | 84.58 ± 0.24 |
| ASN | 82.95 ± 0.37 | 76.05 ± 0.42 | 63.04 ± 0.19 | 59.28 ± 0.31 | 82.35 ± 0.27 | 84.92 ± 0.37 |
| GRADE | 83.19 ± 0.54 | 75.62 ± 0.57 | 62.60 ± 0.47 | 60.73 ± 0.94 | 82.31 ± 0.45 | 84.71 ± 0.51 |
| StrucRW | 89.56 ± 0.80 | 77.21 ± 0.94 | 64.05 ± 0.75 | 59.51 ± 0.44 | 82.78 ± 0.61 | 83.01 ± 0.41 |
| SpecReg | 91.80 ± 0.15 | 76.26 ± 0.05 | 65.74 ± 0.24 | 59.89 ± 0.82 | 83.54 ± 0.42 | 85.24 ± 0.17 |
| A2GCN | 92.01 ± 0.19 | 77.35 ± 0.13 | 65.54 ± 0.26 | 59.42 ± 0.34 | 84.02 ± 0.53 | 85.02 ± 0.26 |
| CMPGNN | **94.02 ± 0.51** | **79.67 ± 0.34** | **66.04 ± 0.61** | **61.43 ± 0.54** | **84.72 ± 0.24** | **86.16 ± 0.21** |

Table 2: AUROC score on Twitch Dataset

| Methods | DE $\to$ EN | FR $\to$ EN | ES $\to$ EN | PT $\to$ EN | DE $\to$ ES | FR $\to$ PT |
|---|---|---|---|---|---|---|
| GCN | 51.95 | 53.52 | 53.92 | 52.68 | 56.17 | 53.56 |
| ASN | 51.00 | 53.03 | 54.64 | 52.13 | 53.57 | 51.39 |
| A2GCN | 53.84 | 54.05 | 53.24 | 53.80 | 59.41 | 53.91 |
| StrucRW | 51.31 | 53.62 | 53.90 | 52.78 | 59.60 | 51.23 |
| SpecReg | 54.67 | 55.50 | 55.43 | 53.12 | 51.04 | 50.89 |
| UDAGCN | 54.70 | 56.52 | 56.58 | 53.01 | 56.48 | 58.27 |
| GRADE | 53.66 | 56.58 | 57.54 | 54.52 | 58.57 | 55.71 |
| CMPGNN | **58.84** | **58.31** | **58.42** | **57.07** | **60.91** | **59.90** |

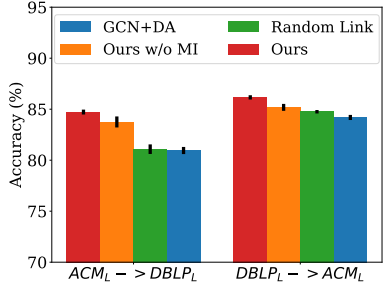

Figure 5: Ablation Study Results

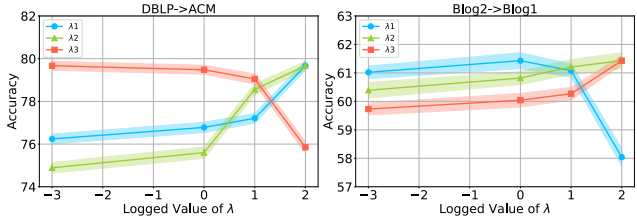

Figure 6: Hyperparameter Sensitivity.

### 5.4 Hyperparameters Analysis

In this section, we further explore how the hyperparameters $\lambda_1$, $\lambda_2$, and $\lambda_3$ affect the model's performance. These hyperparameters govern the contributions of the RA loss $\mathcal{L}_{RA}$, the MI loss $\mathcal{L}_{MI}$, and the entropy loss $\mathcal{L}_{entropy}$ to the final objective. To evaluate the sensitivity of these hyperparameters, we adjust the values of $\lambda_1$, $\lambda_2$, and $\lambda_3$ within the set $\{1e-3, 1e-1, 1, 10, 100\}$, across two experimental setups: ACM to DBLP and Blog1 to Blog2. The outcomes of these experiments are shown in Figure 6. From Figure 6, we can observe: (i) As the value of $\lambda_1$ increases, the model's performance initially improves but may subsequently decrease. This suggests that overemphasizing the contribution of link prediction loss could lead to over-regularization; (ii) As the value of $\lambda_2$ changes, the model's performance does not exhibit drastic fluctuations, indicating that the mutual information loss component is relatively stable.; and (iii) As the value of $\lambda_3$ increases, the model's performance exhibits different trends under the two experimental settings. This indicates that entropy loss is not universally beneficial across all training scenarios.

### 5.5 Probing Further

In the case study, we investigate the characteristics of cross-domain links added during graph domain adaptation.

*5.5.1 Type of Cross-Domain Edges.* One direct question over patterns of added connections is: are they primarily inter-class or intra-class edges? To answer the **RQ2**, we analyze the number of selected edges and the proportion of intra-class cross-domain edges during training in under both homophily setting: ACM to DBLP, and heterophily setting: Blog1 to Blog2, in Figure 7. The homophily ratios for these datasets are presented in Appendix A. Our analysis reveals a consistent pattern across graphs with varying homophily ratios. At the start of training, the link predictor is not yet accurately trained, leading to fluctuations in both the quantity of added edges and the proportion of intra-class edges. This initial stage can introduce noise into the learning of parameters. However, as the training proceeds and the link, the predictor becomes more adept at identifying suitable cross-domain edges, and an increase in the proportion of intra-class edges can be observed. The peak performance of our model also coincides with the peak of the intra-class edge ratio, as highlighted in the figure. This phenomenon reveals the advantage of passing messages from similar source nodes to the target one addressing the shifted domain distribution.

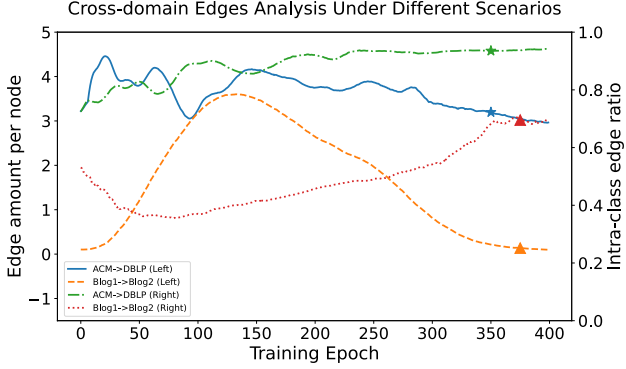

**Figure 7: Evolution of the intra-class edge ratio and the volume of cross-domain edges per node during training for ACM to DBLP and Blog1 to Blog2 scenarios. Solid blue and orange lines denote the edge counts per node, while dashed blue and red lines represent the intra-class edge ratios. Epochs at which optimal model performance is attained are marked.**

*5.5.2 Domain discriminative Information preservation.* Cross-domain propagation through newly inserted cross-domain edges allows nodes in the target domain to receive information from the source domain but meanwhile lead to a loss of domain-discriminative information in the target domain. One key question is whether Mutual Information (MI) loss can help preserve the unique characteristics of the target domain. To address **RQ3**, Figure 8 compares the changes in the embedding similarity between the two domains with and without applying MI loss during the training process. The figure shows that, without MI loss, the embeddings of the target domain become increasingly similar to those of the source domain as training progresses, indicating a loss of discriminative information. In contrast, when MI loss is applied, the embedding similarity between domains remains relatively stable. This suggests that MI loss acts as a regularization mechanism, helping the target domain retain its unique information and mitigating the information-smoothing effect caused by cross-domain propagation.

*5.5.3 Reducing Shifts of Class-wise Distributions.* The introduction of cross-domain edges facilitates message passing from the source to nodes in the target graph. Therefore, one natural question is whether our method can encourage the alignment of conditional embedding distribution $P(H|\hat{Y})$ across domains. To evaluate the alignment ability of our model and answer **RQ4**, we quantify the discrepancy of distributions as the distance between embedding centers of each class, $\mathbb{E}[\mathbf{H}^T \mid y_i] - \mathbb{E}[\mathbf{H}^S \mid y_i]$ for class $i$. Experimental results are visualized in Figure 9. Our model's embedding center shifts are compared with those of UDAGCN [42] and SpecReg [46], which employ adversarial training and discrepancy measurement, respectively. Results validate that our method, through the incorporation of cross-domain edges, is capable of learning embeddings that exhibit minimal center shifts for the majority of classes. A narrower gap in class-wise distribution would help improve the transferability of the source model adapted to the target domain.

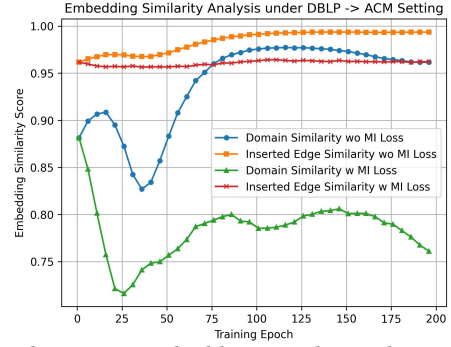

**Figure 8: Changes in embedding similarity during training. The orange and blue lines show the embedding similarity between domains and the node pairs of newly inserted cross-domain edges without Mutual Information Loss, while the red and green lines show the effect when Mutual Information Loss is applied.**

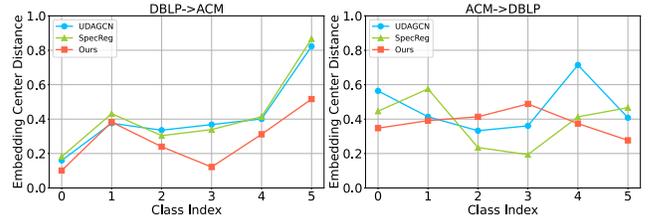

**Figure 9: Embedding Center Distance Among Domains. The gap for each class's embedding center between the source and target domains is calculated for two experimental settings: ACM to DBLP and DBLP to ACM.**

## 6 Conclusion

In this work, we introduce a novel approach to graph domain adaptation by incorporating cross-domain edges to enhance message passing and reduce distribution discrepancies between source and target graphs. Our framework, Cross-domain Message Passing Graph Neural Network (CMPGNN), uses an edge predictor to create cross-domain edges that facilitate knowledge transfer and align target graph embeddings with those of the source domain. Additionally, a mutual information-based loss helps maintain discriminative information for target nodes. Experiments on various datasets demonstrate CMPGNN's effectiveness in aligning conditional distributions, even with label shifts.

## 7 Acknowledgments

This material is based upon work supported by the U.S. Department of Homeland Security under Grant Award Number 17STCIN00001-05-00. The views and conclusions contained in this document are those of the authors and should not be interpreted as necessarily representing the official policies, either expressed or implied, of the U.S. Department of Homeland Security.

## Ethical Considerations

We have thoroughly evaluated the ethical implications and societal impact of our research. We are confident that our methodology and results do not pose notable risks concerning fairness, privacy, security, or safety. We remain committed to proactively addressing any ethical concerns that may emerge as this work progresses.

Table 3: Summary of Datasets Used in Experiments

| Dataset | #Domains | #Nodes | #Edges | #Homo Ratio | #Attributes | #Labels |
|---|---|---|---|---|---|---|
| Citation | ACM | 7,410 | 11,135 | 0.83 | 7,537 | 6 |
|  | DBLP | 5,578 | 7,341 | 0.96 |  |  |
| Blog | Blog1 | 2,300 | 66,942 | 0.39 | 500 | 6 |
|  | Blog2 | 2,896 | 107,672 | 0.40 |  |  |
| Citation (Large) | $ACM_L$ | 38,527 | 74,510 | 0.81 | 2,000 | 3 |
|  | $DBLP_L$ | 41,943 | 25,112 | 0.93 |  |  |
| Twitch | DE | 9,498 | 153,138 | 0.63 | 3,170 | 2 |
|  | FR | 6,549 | 112,666 | 0.56 |  |  |
|  | EN | 7,126 | 35,324 | 0.56 |  |  |
|  | PT | 1,912 | 31,299 | 0.57 |  |  |
|  | ES | 4,648 | 59,382 | 0.58 |  |  |

## A Benchmark Datasets and Baseline Implementation

### A.1 Benchmark Introduction

Here is a detailed introduction to the benchmark we used in the experiment.

- **ACM and DBLP:** ACM and DBLP are two citation networks from ArnetMiner [33, 42]. Nodes represent papers categorized into topics such as "Database", "Data Mining", "Artificial Intelligence", "Computer Vision", "Information Security", and "High-Performance Computing".
- **Blog Datasets:** Originating from BlogCatalog [47], this dataset features two social networks. Nodes symbolize bloggers, edges indicate friendships, and labels correspond to the bloggers' community affiliations. Attributes, reduced via PCA [40] to 500 dimensions, encapsulate keywords from the bloggers' profiles.
- **$ACM_L$ and $DBLP_L$:** Current datasets used in graph domain adaptation tasks are usually tiny, containing only a few thousand nodes. To address this limitation, we construct larger citation networks from extended versions of ACM and DBLP datasets, denoted as $ACM_L$ and $DBLP_L$ respectively. These span broader publication years, with $ACM_L$ covering 1984 to 2014 and $DBLP_L$ post-2016, ensuring no overlap with nodes in the smaller versions. Papers are classified into "Data Mining", "Computer Vision", and "Natural Language Processing", and represented by a 2,000-dimensional bag-of-words abstract attribute.
- **Twitch** [19]: Gamer networks from different regions. The nodes represent users and connections denote friendships. Node features include users' preferred games, geographical location, and streaming habits. Users are grouped into two categories based on their use of explicit language. The label shift issue is significant in this dataset.

Table 3 gives detailed statistics of the datasets employed in our experiments.

### A.2 Baseline Models and Implementation Details

For a comprehensive evaluation, we benchmark our proposed CMPGNN model against the following baseline methods, including the latest state-of-the-art domain adaptation techniques in graph domains:

- **GCN** [15]: It is a foundational deep convolutional neural network for graph data, which employs low-frequency filtering to aggregate embeddings through the graph's structure.
- **GRADE** [41]: It utilizes Graph Subtree Discrepancy to quantify distribution shifts between source and target graphs, eschewing adversarial training approaches.
- **UDAGCN** [42]: It integrates local and global structural patterns with an inter-graph attention mechanism during domain adversarial training.
- **ASN** [50]: It utilizes domain-specific encoders to preserve both domain-private and domain-invariant features.
- **StrucRW** [21]: It adapts source graph edge weights to mirror target graph connectivity trends, addressing domain connection pattern shifts.
- **SpecReg** [46]: It employs spectral domain regularization to align embeddings across source and target domains.
- **A2GCN** [19]: It differentiates propagation depths across domains and mitigates the influence of the source domain on the target domain.

Our implementations are based on the PyTorch framework [27], utilizing the PyTorch Geometric (PyG) library [6] for graph data handling. Each model uses a standardized graph convolutional encoder architecture (Input × 128 × 64) to process feature information. The Adam optimizer [14] is utilized for model training across all baselines. Hyperparameters for each model were fine-tuned using a grid search approach. Shared hyperparameters like the learning rate were kept consistent during optimization, while model-specific hyperparameters were tuned according to the guidelines outlined in their respective publications.

## B Training Algorithm and Time Complexity Analysis

In this section, we give the whole training algorithm and time complexity analysis of the proposed CMPGNN model.

### B.1 Training Algorithm

The full pipeline of our framework can be summarized in Algorithm 1. Before training starts, we select the candidate set for each node in the target domain in line 1. Inside each optimization step, we first obtain node representations for each graph using the encoder in line 3. Then we calculate the mutual information loss $\mathcal{L}_{MI}$ in the embedding space in Line 4. From line 5 to line 8, we feed the embedding into the classifier $f_c$ and calculate the classification loss $\mathcal{L}_{clf}$ and $\mathcal{L}_{entropy}$ in the predicted label space. In line 8 and line 9, we feed the $H^C$ to the link predictor and calculate the link predictor loss $\mathcal{L}_{RA}$. In line 9, we train the model with the full adaptation objective and update the parameter. Finally, in Section 4.2.2, we select and insert cross-domain edges into the combined graph $\mathcal{G}^C$

### B.2 Time Complexity

For time complexity analysis, consider the source graph $\mathcal{G}^S$ with $N^S$ nodes, $M^S$ edges, and the target graph $\mathcal{G}^T$ with $N^T$ nodes and $M^T$ edges. Selecting the edge candidate set before training requires calculating the embedding similarity among each cross-domain node pair, which involves $O(N^S N^T)$ time complexity. During each training epoch, assuming the graph is sparse, the graph convolution operation on the source and target graph needs $O(M^S + M^T)$ time complexity. The link predictor loss is applied to each edge in

Table 4: Classification Accuracy Under Semi-Supervised-Setting

| Methods | ACM → DBLP | DBLP → ACM | Blog1 → Blog2 | Blog2 → Blog1 | $ACM_L$ → $DBLP_L$ | $DBLP_L$ → $ACM_L$ |
|---|---|---|---|---|---|---|
| GCN | 76.18 ± 0.34 | 73.61 ± 0.54 | 51.62 ± 0.51 | 50.95 ± 0.43 | 79.65 ± 0.37 | 81.31 ± 0.47 |
| UDAGCN | 82.05± 0.45 | 75.49 ± 0.31 | 56.54 ± 0.32 | 57.23 ± 0.19 | 82.13 ± 0.21 | 83.61 ± 0.17 |
| ASN | 83.61 ± 0.24 | 76.27 ± 0.19 | 58.01 ± 0.24 | 56.93 ± 0.20 | 81.98 ± 0.29 | 83.80 ± 0.25 |
| GRADE | 84.02 ± 0.37 | 75.24 ± 0.27 | 57.66 ± 0.36 | 57.52 ± 0.17 | 82.78 ± 0.31 | 83.07 ± 0.43 |
| StrucRW | 88.90 ± 0.24 | 76.83 ± 0.38 | 59.25 ± 0.34 | 58.68 ± 0.16 | 82.04 ± 0.16 | 82.95 ± 0.25 |
| SpecReg | 90.91 ± 0.31 | 77.35 ± 0.47 | 60.01 ± 0.38 | 58.45 ± 0.40 | 83.25 ± 0.37 | 85.08 ± 0.18 |
| A2GCN | 91.81 ± 0.24 | 77.28 ± 0.16 | 61.63 ± 0.51 | 57.12 ± 0.21 | 83.86 ± 0.24 | 85.04 ± 0.16 |
| **CMPGNN** | **94.81 ± 0.20** | **79.65 ± 0.15** | **62.21 ± 0.24** | **58.96 ± 0.13** | **84.58 ± 0.27** | **85.94 ± 0.17** |

Table 5: Label-shift among Domains

| Class | 0 | 1 | 2 | 3 | 4 | 5 |
|---|---|---|---|---|---|---|
| ACM | −0.18 | 0.50 | −0.59 | −0.40 | 4.91 | 2.30 |
| DBLP | 0.22 | −0.33 | 1.45 | 0.67 | −0.83 | −0.70 |

**Algorithm 1** Training Algorithm for the Cross-domain Message Passing Graph Neural Network

**Require:** source and target graph $\mathcal{G}^S$, $\mathcal{G}^T$; initial combined graph $\mathcal{G}^C$; pre-trained encoder $f_E$; epoch num $n$

1: Pre-select the edge candidate set for each node in the target graph, according to Section 4.2.2
2: **while** epoch < $n$ **do**
3: 　Feed graphs into encoder $f_E$ and get embedding $H^S, H^T, H^C$ on different graphs
4: 　Calculate the mutual information loss $\mathcal{L}_{MI}$ between combined and target graph according to Equation 6
5: 　Feed $H^S$, and $H_C^T$ to the classifier $f_{cls}$
6: 　Calculate the classification loss $\mathcal{L}_{clf}$ on the source graph according to Equation 7
7: 　Calculate the entropy loss $\mathcal{L}_{clf}$ on the combined graph according to Equation 8
8: 　Feed the embedding $H^C$ on combined graph to the link predictor and get embedding $Z_C^S$, and $Z_C^T$
9: 　Calculate the link prediction loss according to Equation 3
10: 　Get the final objective and update parameters according to Equation 9
11: 　Modify the combined graph by adding cross-domain edges with adaptive weights, described in Section 4.2.2
12: **end while**

the source and target domain, which requires $O(M^S + M^T)$ time complexity. Furthermore, the mutual information loss is computed for each node within the target domain by contrasting the augmented embedding against the negative samples derived from the original target graph, introducing an additional time complexity of $O(N^T)$. Finally, the process of link insertion requires $O(N^T)$ time complexity. Overall, the time complexity for the pre-selection phase is $O(N^S N^T)$, and the time complexity for the training procedure is $O(M^S + M^T + N^T)$.

## C Experiment result under semi-supervised setting

In this section, we provide more results of the evaluation of our proposed CMPGNN model against baseline models under semi-supervised setting. The experiment results are shown in Table 4.